\pdfoutput=1
\documentclass[11pt]{article}

\usepackage[preprint]{acl}

\usepackage{times}
\usepackage{latexsym}

\usepackage{epstopdf}
\usepackage{algorithm}
\usepackage{algorithmic}
\usepackage{amsmath}
\usepackage{amsthm}
\usepackage{graphicx}
\usepackage{latexsym}
\usepackage{multirow}
\usepackage{multicol}
\usepackage{diagbox}
\usepackage{makecell}
\usepackage{graphics}
\usepackage{amsfonts,amssymb} 
\usepackage{booktabs}
\usepackage{tabularx}
\usepackage{float}
\usepackage{enumitem}
\usepackage{subfigure}
\newtheorem{theorem}{Theorem}[section]

\usepackage[T1]{fontenc}
\usepackage[utf8]{inputenc}

\usepackage{microtype}

\usepackage{inconsolata}

\usepackage{graphicx}

\title{OPT-Tree: Speculative Decoding with Adaptive Draft Tree Structure}

\author{
 \textbf{Jikai Wang\textsuperscript{1}\footnotemark[1]},
 \textbf{Yi Su\textsuperscript{1}\thanks{Equal contribution.}},
 \textbf{Juntao Li\textsuperscript{1}\thanks{Corresponding author.}},
\\
 \textbf{Qingrong Xia\textsuperscript{2}},
 \textbf{Zi Ye\textsuperscript{2}},
 \textbf{Xinyu Duan\textsuperscript{2}},
 \textbf{Zhefeng Wang\textsuperscript{2}},
 \textbf{Min Zhang\textsuperscript{1}},
\\
 \textsuperscript{1}Institute of Computer Science and Technology, Soochow University, China\\
 \textsuperscript{2}Huawei Cloud
\\
    \href{mailto:risus254@gmail.com}{risus254@gmail.com},
    \href{mailto:yisunlp@outlook.com}{yisunlp@outlook.com}\\
    \href{mailto:ljt@suda.edu.cn}{ljt@suda.edu.cn}
 }

\begin{document}
\maketitle
\begin{abstract}
Autoregressive language models demonstrate excellent performance in various scenarios.
However, the inference efficiency is limited by its one-step-one-word generation mode, which has become a pressing problem recently as the models become increasingly larger.
Speculative decoding employs a "draft and then verify" mechanism to allow multiple tokens to be generated in one step, realizing lossless acceleration.
Existing methods mainly adopt fixed heuristic draft structures, which fail to adapt to different situations to maximize the acceptance length during verification.
To alleviate this dilemma, we proposed OPT-Tree, an algorithm to construct adaptive and scalable draft trees.
It searches the optimal tree structure that maximizes the mathematical expectation of the acceptance length in each decoding step.
Experimental results reveal that OPT-Tree outperforms the existing draft structures and achieves a speed-up ratio of up to 3.2 compared with autoregressive decoding. 
If the draft model is powerful enough and the node budget is sufficient, it can generate more than ten tokens in a single step.
Our code is available at \href{https://github.com/Jikai0Wang/OPT-Tree}{https://github.com/Jikai0Wang/OPT-Tree}.

\end{abstract}

\section{Introduction}
Large language models (LLMs) \citep{black-etal-2022-gpt,touvron2023llama,achiam2023gpt,zheng2024judging} have achieved remarkable performance in various NLP scenarios. 
As models grow in size and complexity, the computational demands for inference increase significantly.
Therefore, it is becoming increasingly important to accelerate decoding to save computing overhead.

Autoregressive models \citep{black-etal-2022-gpt,zhang2022opt,touvron2023llama} usually generate one token in one decoding step, leading to limited decoding efficiency.
In recent work, speculative decoding \citep{leviathan2023fast,he2023rest,fu2024break,cai2024medusa,li2024eagle} has shown great potential for lossless accelerated decoding.
It applies a "draft and then verify" mechanism to maintain the original output distribution of the target model to be accelerated.
Drafting is performed by a less-overhead drafting model.
The generated draft is verified in parallel by the target model to generate multiple tokens in one decoding step, bringing promising acceleration.

\begin{figure*}[t]
    \centering
    \includegraphics[width=\textwidth]{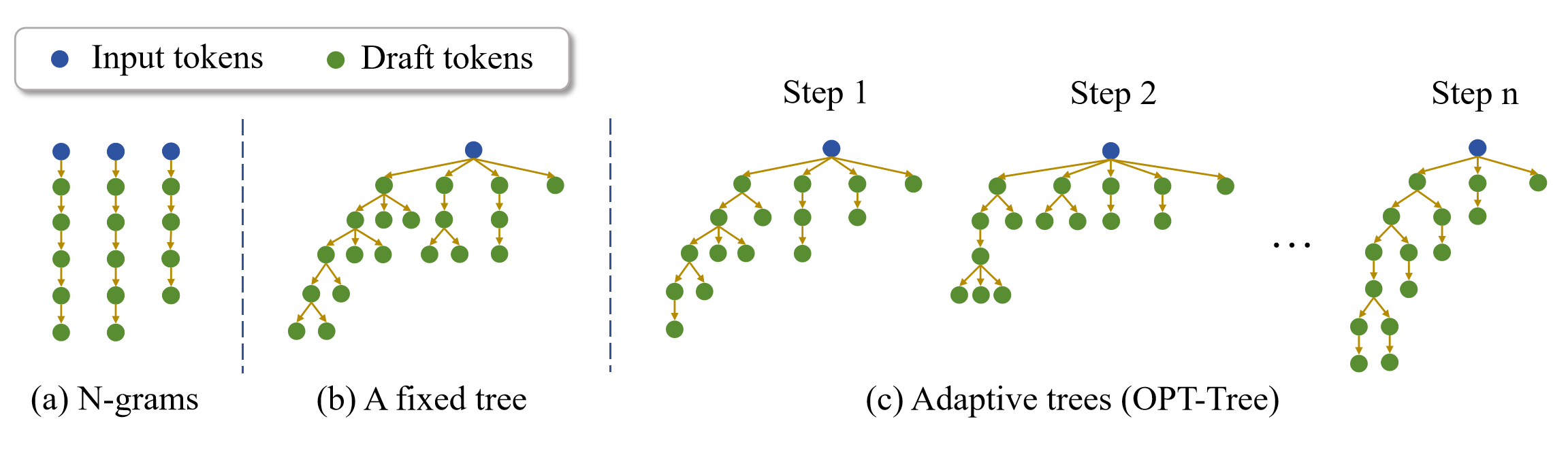}
    \caption{Draft structures used in speculative decoding. Nodes in the same layer share the same position index. OPT-Tree various in each decoding step to achieve a larger acceptance length.}
    \label{fig:intro}
\end{figure*}

Existing work like EAGLE \cite{li2024eagle} has proposed methods for training small but effective draft models.
To the best of our knowledge, previous work mainly adopts drafts with structures of Sequences or fixed trees.
However, we argue that neither of them is the optimal draft structure under a limited node budget.
Sequence-structured drafts \citep{stern2018blockwise,leviathan2023fast,xia2023speculative,yang2023predictive,zhang2023draft,fu2024break} contain redundant nodes. For example, "A-B-C-D-E" and "A-B-C-F-G" have the same prefix "A-B-C", which is calculated twice during verification. Therefore, there are only 7 valid tokens among the 10 nodes of these two sequences.
Drafts with tree structure \citep{he2023rest,cai2024medusa,li2024eagle,jeon2024recursive,chen2024sequoia} solved this problem. 
The same token can only appear once in the same tree layer.
A corresponding tree attention mask is designed for parallel verification.
The specific structure of the tree is usually heuristic and remains constant.
However, given a node budget, the best structure that maximizes the acceptance length during verification would change according to different inputs in each decoding step.

This paper proposes an adaptive and scalable tree structure called OPT-Tree.
It can be applied to any autoregressive draft model.
As is shown in Figure \ref{fig:intro}, the tree structure adaptively changes in each decoding step to maximize the mathematical expectation of the acceptance length. 
We apply a greedy algorithm to construct an OPT-Tree in each step.
Details are elaborated in Section \ref{sec:method}.
We conduct comprehensive experiments in Section \ref{sec:experiments} to evaluate the effectiveness of OPT-Tree.
Experimental results demonstrate that OPT-Tree outperforms the baselines and can be up to 3.2 times faster than vanilla autoregressive decoding.
% Section \ref{sec:case} shows an intuitive case in which over 9 tokens are generated in one step on average.
The mathematical expectation of the acceptance length is generally positively correlated with the actual acceptance length in practice.
Moreover, OPT-Tree performs well when the tree size scales up.
Using LLaMA-2-7B as the draft model, LLaMA-2-7B can generate 10 tokens in a single decoding step with OPT-Tree when the number of nodes is over 500, which indicates its great potential for adapting to more powerful computation resources and more effective draft models in the future.

\section{Preliminaries}
We provide the necessary definitions in this section.

\noindent{\textbf{Inference.}} After inputting $\textbf{x}=(x_{1},x_{2},...,x_{l})$, where $l$ is the current sequence length, the target model $M$ and the drafting model $M_{d}$ return the next word distribution $p(y^{l+1}|x_{1},x_{2},...,x_{l})$ and $p_{d}(\hat{y}^{l+1}|x_{1},x_{2},...,x_{l})$ respectively, where $y^{l+1}$ and $\hat{y}^{l+1}$ are the sampled next words.

\noindent{\textbf{Speculative Decoding.}} In speculative decoding with tree-structured draft, $M_{d}$ first infers $d$ steps to generate a draft tree $T$ of depth $d$ and then $M$ verify the draft. 
The verification depends on the sampling method.
For greedy sampling, the ground truth is the sequence of tokens with the highest probability for each position output by $M$. 
For all branches in the tree that contain the root node, the longest branch with the same prefix as the ground truth is accepted.
Therefore, multiple tokens can be generated in one decoding step while ensuring that the generated sequences are consistent with the original ones.

\section{OPT-Tree}
\label{sec:method}
This section introduces OPT-Tree, an algorithm for constructing our defined optimal draft tree structure for any input sequence in speculative decoding with autoregressive draft models.
%Note that, we only consider greedy sampling in this section for the sake of brevity.

% First we give some definitions.
% After inputting $\textbf{x}=(x_{1},x_{2},...,x_{l})$, where $l$ is the current sequence length, the target model $M$ and the drafting model $M_{d}$ return the next word distribution $p(y^{l+1}|x_{1},x_{2},...,x_{l})$ and $p_{d}(\hat{y}^{l+1}|x_{1},x_{2},...,x_{l})$ respectively, where $y^{l+1}$ and $\hat{y}^{l+1}$ are the sampled next words.
% In speculative decoding, $M_{d}$ first infers $d$ steps to generate a draft tree $T$ of depth $d$ and then $M$ verify the draft. Multiple tokens can be generated in one decoding step.
Draft tree $T$ is defined as follows:
\begin{equation}
\begin{aligned}
    &T=(\mathbb{V},\mathbb{E}) \\
    \mathbb{V}=&\bigcup_{i=l+1}^{l+d}\bigcup_{j=1}^{n_{i}}\left\{ (\hat{y}^{i}_{j},\hat{p}^{i}_{j})\right\},
\end{aligned}
\end{equation}
where $\mathbb{V}$ and $\mathbb{E}$ is the set of all nodes and edges.
% \begin{equation}
%     \mathbb{V}=\bigcup_{i=l+1}^{l+d}\bigcup_{j=1}^{n_{i}}\left\{ (\hat{y}^{i}_{j},\hat{p}^{i}_{j})\right\},
% \end{equation}
$n_{i}$ represents the number of sampled tokens in the $i_{th}$ layer of $T$.
$\hat{p}^{i}_{j}$ is calculated by:
\begin{equation}
    \hat{p}^{i}_{j}=\prod_{\hat{y}\in \mathbb{P}(\hat{y}_{j}^{i})}p_{d}(\hat{y}),
\end{equation}
where $\mathbb{P}(\hat{y}_{j}^{i})$ is the set of all parent nodes of $\hat{y}_{j}^{i}$ (including itself). 
$\hat{p}^{i}_{j}$ of the root node is regarded as positive infinity.
% Theorem \ref{the:1} can be obtained from the above conditions. 
For each node in $T$, if it has $k$ children, they are $k$ tokens greedily sampled according to $p_{d}$ from its subsequent token distribution. The purpose of calculating $\hat{p}$ is to simplify subsequent operations. 
\begin{theorem}
    \label{the:1}
    For any two nodes $v_{i}$ and $v_{j}$ in the tree, if $v_{i}$ is a node in the subtree of $v_{j}$, then $\hat{p}$ of $v_{i}$ is less than $\hat{p}$ of $v_{j}$.
\end{theorem}
\begin{figure}[t]
    \centering
    \includegraphics[width=\columnwidth]{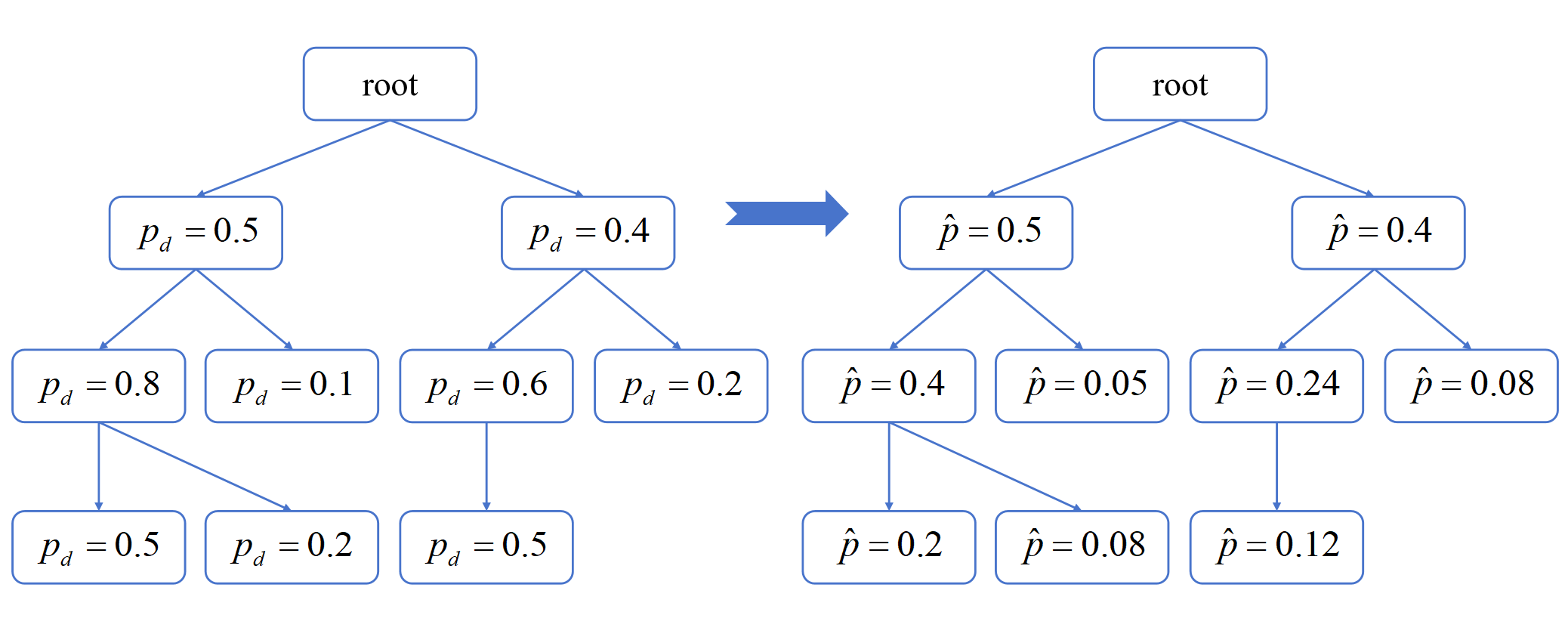}
    \caption{An example of a draft tree containing $\hat{p}$ in each node. The value of $E(A)$ is 2.07.}
    \label{fig:tree}
\end{figure}

Considering a certain step in speculative decoding whose input is $\textbf{x}$, the draft model $M_{d}$ generates a draft tree based on $\textbf{x}$ and the given tree structure $T$. Then, the target model inputs the draft tree and the corresponding tree attention mask and returns the next tokens of each token in $T$. 
We get the longest accepted candidate with length $A$ by comparing the next tokens and the draft tree.
% maximize the acceptance length $A$ of the current draft during verification. 
% $T_{opt}$ is related to the output of the target model. However, we need to determine tree structure before verification in actual scenarios when $A$ is unobtainable. But we can predict $A$ based on the distribution information of output by $M_{d}$. We define $T_{opt}$ as the tree structure that 
Given $M$, $M_{d}$ and $n$, for input $\textbf{x}$, an optimal tree structure $T_{opt}$ should maximize the mathematical expectation of the acceptance length $E(A)$. Note that $T_{opt}$ changes as the input changes.
Since the optimization goal of the draft model is to make its output distribution close to the target model distribution, for each node, $\hat{p}$ will be positively related to its probability of being accepted during verification when using an effective draft model for speculative decoding. 
Therefore, $E(A)$ can be approximately calculated by $\hat{p}$:
\begin{equation}
    \begin{aligned}
    E(A)&=\sum_{(\hat{y}^{i}_{j},\hat{p}^{i}_{j})\in T} \prod_{\hat{y}\in \mathbb{P}(\hat{y}_{j}^{i})}p_{d}(\hat{y}) \\
    &=\sum_{(\hat{y}^{i}_{j},\hat{p}^{i}_{j})\in T}\hat{p}^{i}_{j}.
    \end{aligned}
\end{equation}
Figure \ref{fig:tree} shows a simple example of calculating $\hat{p}$ and $E(A)$.
$E(A)$ should positively correlate with the acceptance length.
We discuss their correlation in Section \ref{sec:e}.

% As shown in Figure 1, $T$ satisfies Theorem \ref{the:1}.

We use $E_{sub}(T,n)$ to represent the maximum value of $E(A)$ for all subtrees of $T$ that contain the root node and have n nodes.
Note that the root node is not considered when calculating node trees and mathematical expectations.
% Figure \ref{fig:e} shows an example of calculating $E(A)$.
% We conduct an experiment in Section \ref{sec:similarity} to study the similarity between the two trees.

Then, we propose Algorithm \ref{alg:draft} to construct $T_{opt}$ during the drafting phase for each decoding step. We initialize $T$ with a root node. At each drafting step, we greedily sample n tokens with the largest $\hat{p}$ in the next token distributions of nodes in the last layer of $T$ to construct the next layer. $T$ has $d*n$ nodes at this time. Finally, we select the $n$ nodes in $T$ with the largest $p$. It is easy to prove that these $n$ nodes are a subtree of $T$, which contains the root node:
\begin{proof}
(1) If these nodes can not form a tree with the root, there is at least one node $v_{i}$ whose parent node $v_{j}$ is not among these nodes. (2) According to Theorem \ref{the:1}, $\hat{p}$ of $v_{j}$ is larger than $\hat{p}$ of $v_{i}$. Therefore, $v_{j}$ is also selected. (1) and (2) are contradictory, so these nodes must be able to form a subtree of $T$ containing the root node.
\end{proof}

% According to Theorem 3.2, this tree is the desired $T_{opt}$.
\renewcommand{\algorithmicrequire}{ \textbf{Input:}}
\renewcommand{\algorithmicensure}{ \textbf{Output:}}
\begin{algorithm}
\caption{Construct an OPT-Tree $T_{opt}$}
\label{alg:draft}
\begin{algorithmic}
\REQUIRE ~~
Input sequence $\textbf{x}=(x_{1},x_{2},...,x_{l})$, draft model $M_{d}$, number of nodes $n$, threshold $\delta$.
\ENSURE ~~A draft tree $T_{opt}$.
\STATE Initialize a tree $T$ with root node $x_{l}$
\STATE $E \leftarrow 0$
\STATE Output distribution $P_{d}(T)\leftarrow M_{d}(T)$
\STATE  $T \leftarrow topk(P_{d}(T),n)$
% \STATE $//$ Greedily sample n tokens in the next token distributions of nodes in the last layer with the largest $\hat{p}$ to construct the next layer
\WHILE {Depth of tree $D(T)<n$ and $E_{sub}(T,n)-E>\delta$}
\STATE $//$ Drafting step
\STATE $E \leftarrow E_{sub}(T,n)$
\STATE Output distribution $P_{d}(T)\leftarrow M_{d}(T)$
% \STATE $T \leftarrow UPDATE(T,P_{d}(T))$
\STATE  $T \leftarrow topk(P_{d}(T),n)$
\ENDWHILE
% \STATE $//$ Get a tree $T$ with $d*n$ nodes (excluding the root node)
%从n*dt个节点中取出top n
\STATE $T_{opt} \leftarrow$ Select the $n$ nodes with the largest $\hat{p}$ from $T$
\end{algorithmic}
\end{algorithm}

\begin{theorem}
    \label{the:2}
    As the drafting step increases, $E_{sub}(T,n)$ is monotonic non-decreasing.
\end{theorem}

\begin{algorithm}
\caption{Speculative Decoding with Adaptive Draft Tree Structure}
\label{alg:decoding}
\begin{algorithmic}
\REQUIRE ~~
Input sequence $\textbf{x}=(x_{1},x_{2},...,x_{l})$, target model $M$, draft model $M_{d}$, number of nodes $n$, threshold $\delta$.
%\ENSURE ~~An accepted sequence of tokens $(y^{l+1},y^{l+2},...,y^{l+A})$ with length $A$.
\ENSURE ~~New input sequence $\textbf{x}^{\prime}=(x_{1},x_{2},...,$\\$x_{l+A})$
\STATE $T_{opt} \leftarrow$ Construct the draft tree with $n$ nodes
\STATE $mask \leftarrow$ Compute the corresponding tree attention mask
\STATE $P \leftarrow M(T_{opt},mask)$ 
\STATE $(y^{l+1},y^{l+2},...,y^{l+A})\leftarrow Verify(T_{opt},P)$ 
\STATE $//$ Find the longest accepted candidate. If a sequence of length $A-1$ successfully hits, its next word will also be accepted. So, the total acceptance length is $A$.
\STATE $\textbf{x}^{\prime} \leftarrow Concat(\textbf{x},(y^{l+1},y^{l+2},...,y^{l+A}))$
\end{algorithmic}
\end{algorithm}
\begin{table*}
\centering
\small
\renewcommand{\arraystretch}{1.45}
\resizebox{\textwidth}{!}{
\begin{tabular}{c|c|c|ccc||c|c|c|ccc} 
\toprule
\textbf{$M$}           & \textbf{$M_{d}$ }                     & Tree     & MAL    & Tokens/s & Speedup & \textbf{$M$}          & \textbf{$M_{d}$ }                     & Tree     & MAL    & Tokens/s & Speedup  \\ 
\midrule
\multirow{10}{*}{\begin{tabular}[c]{@{}c@{}}LLaMA-2\\-7B\end{tabular}} & None                    & -        & 1.00 & 51.89    & 1.00    & \multirow{10}{*}{\begin{tabular}[c]{@{}c@{}}LLaMA-2\\-13B\end{tabular}} & None                   & -        & 1.00 &    26.79      & 1.00     \\ 
\cline{2-6}\cline{8-12}
  & \multirow{3}{*}{L-68M}  & Binary   &   2.12   &      68.58    &     1.32    &                                                                        & \multirow{3}{*}{L-68M} & Binary   &    2.05  &      40.24    &     1.50     \\
                                                                      &                         & EAGLE    &  2.47    &      77.06    &    1.49     &                                                                        &                        & EAGLE    &    2.42  &     46.82     &     1.75     \\
                                                                      &                         & OPT-Tree &  \textbf{2.58}     &     \textbf{87.57}     &    \textbf{1.69}      &                                                                        &                        & OPT-Tree &   \textbf{2.58}   &       \textbf{48.10}   &    \textbf{1.80}      \\ 
\cline{2-6}\cline{8-12}
                                                                      & \multirow{3}{*}{L-1B}   & Binary   &  3.95    &      46.10    &    0.89    &                                                                        & \multirow{3}{*}{L-1B}  & Binary   &  3.95    &      37.37    &    1.39      \\
                                                                      &                         & EAGLE    &   4.23   &    47.74      &     0.92    &                                                                        &                        & EAGLE    &   4.25   &   40.12       &    1.50      \\
                                                                      &                         & OPT-Tree &     \textbf{4.88} &    \textbf{52.48}     &   \textbf{1.01}     &                                                                        &                        & OPT-Tree &   \textbf{5.20}   &        \textbf{43.40}  &    \textbf{1.62}      \\ 
\cline{2-6}\cline{8-12}
                                                                      & \multirow{3}{*}{~EAGLE} & Binary   &   3.40   &     107.91      &     2.08    &                                                                        & \multirow{3}{*}{EAGLE} & Binary   &  3.54    &     66.24    &    2.47      \\
                                                                      &                         & EAGLE    &   3.73   &     130.50     &    2.51     &                                                                        &                        & EAGLE    &   3.80   &      73.97    &    2.76      \\
                                                                      &                         & OPT-Tree &   \textbf{4.36}   &    \textbf{132.75}      &    \textbf{2.56}     &                                                                        &                        & OPT-Tree &  \textbf{4.35}   &      \textbf{76.61}    &    \textbf{2.86}      \\ 
\midrule
\multirow{7}{*}{\begin{tabular}[c]{@{}c@{}}LLaMA-2\\-70B\end{tabular}} & None                    & -        &  1.00    &     6.29     &    1.00     & \multirow{7}{*}{\begin{tabular}[c]{@{}c@{}}Vicuna\\-33B\end{tabular}}  & None                   & -        &   1.00   &     11.25     &       1.00   \\ 
\cline{2-6}\cline{8-12}
                                                                      & \multirow{3}{*}{L-7B}   & Binary   &   4.84   &   11.05       &     1.76    &                                                                        & \multirow{3}{*}{V-7B}  & Binary   &  4.41    &       12.49   &  1.11        \\
                                                                      &                         & EAGLE    &   4.97   &      11.35    &  1.80       &                                                                        &                        & EAGLE    &  4.64    &  12.99        &  1.15        \\
                                                                      &                         & OPT-Tree &   \textbf{7.74}   &     \textbf{11.65}     &     \textbf{1.85}     &                                                                       &                        & OPT-Tree &    \textbf{6.51}  &       \textbf{13.74}   &  \textbf{1.22}        \\ 
\cline{2-6}\cline{8-12}
                                                                      & \multirow{3}{*}{EAGLE}  & Binary   &   3.39   &      17.02    &    2.71     &                                                                        & \multirow{3}{*}{EAGLE} & Binary   &   2.35   &   
                                                                      21.13&   1.88      \\
                                                                      &                         & EAGLE    &   3.67   &     18.81     &    2.99      &                                                                        &                        & EAGLE    &2.69
                                                                      &     24.92     &     2.21    \\
                                                                      &                         & OPT-Tree &  \textbf
                                                                      {4.06}   &   \textbf{19.21}       &   \textbf{3.05}      &                                                                        &                        & OPT-Tree &   \textbf{3.06}   &     \textbf{25.17}     &     \textbf{2.24}     \\
\bottomrule
\end{tabular}
}
\caption{\label{tab:main} Experimental results on MT-Bench. $M_{d}$ being None represents vanilla autoregressive decoding. "L" and "V" in $M_{d}$ column represent "LLaMA-2" and "Vicuna". "MAL" indicates "Mean Acceptance Length".}
\end{table*}

% \begin{table}[t]
% \centering
% \small
% \resizebox{\textwidth}{!}{
% \begin{tabular}{cccccc||cccccc}
% \toprule
% M& MD & Tree & A & Tokens/s & Speedup & M & MD & Tree & A & Tokens/s & Speedup\\
%  \hline
%  \multirow{10}{*}{LLaMA-7B} & None & - & 1 & 51.89 & 1 &
%  \multirow{10}{*}{LLaMA-13B} & None & - & 1 &  & 1\\
% \hline
%  &\multirow{3}{*}{L-68M} & Binary & 0 & 0 & 0 &  &\multirow{3}{*}{L-68M} & Binary &  &  &  &\\
%  & & EAGLE &  &  &  & 
%   &  & EAGLE &  &  & \\
%  & & OPT-Tree &  &  &  & 
%    &  & OPT-Tree &  &  &\\
% \hline
%  &\multirow{3}{*}{L-68M} & Binary & 0 & 0 & 0 &  &\multirow{3}{*}{L-68M} & Binary &  &  &  &\\
%  & & EAGLE &  &  &  & 
%   &  & EAGLE &  &  & \\
%  & & OPT-Tree &  &  &  & 
%    &  & OPT-Tree &  &  &\\
% \hline
%  &\multirow{3}{*}{L-68M} & Binary & 0 & 0 & 0 &  &\multirow{3}{*}{L-68M} & Binary &  &  &  &\\
%  & & EAGLE &  &  &  & 
%   &  & EAGLE &  &  & \\
%  & & OPT-Tree &  &  &  & 
%    &  & OPT-Tree &  &  &\\
% \bottomrule
% \end{tabular}
% }
% \caption{\label{tab:main} 1}
% \end{table}

According to Theorem \ref{the:2}, we can get the desired $T_{opt}$ in theory by stopping drafting when $E(T)$ no longer increases. However, the draft model brings additional overhead to the practice. For autoregressive draft models, the drafting overhead is proportional to the depth of the draft tree. 

Taking this into consideration, we introduce a threshold $\delta$ when setting the conditions for terminating drafting. The value of $\delta$ should be controlled between $\mu$ and 1, where $\mu$ is the time of one drafting step divided by the time of one decoding step.

A complete decoding step of $M$ is shown in Algorithm \ref{alg:decoding}.
In practice, both $M$ and $M_{d}$ use key and value cache to calculate attention. Thus, the actual input length of each drafting step is $n$, which avoids computational bottlenecks in the inference of draft model under larger budgets of tree size.

\section{Experiments}
\label{sec:experiments}
\subsection{Main Results}
\label{sec:main}

\noindent{\textbf{Setup.}} We adopt LLaMA-2-7B, LLaMA-2-13B, LLaMA-2-70B \citep{touvron2023llama} and Vicuna-33B \citep{zheng2024judging} as target models to verify the effectiveness of OPT-Tree. 
We use a single GeForce RTX 4090 GPU for LLaMA-2-7B, a single L20 GPU for LLaMA-2-13B and 4 A100-PCIE-40GB GPUs for LLaMA-2-70B and Vicuna-33B.
We choose one or two smaller models in the same version as the draft model for each target model. Moreover, we adopt a corresponding EAGLE draft model for each target model. 
The temperature is set to zero.
EAGLE \citep{li2024eagle} is an effective speculation decoding method that trains additional autoregressive heads as draft models. 
It uses a well-designed heuristic draft tree structure with 25 nodes.
In our experiments, we regard it as the EAGLE draft tree.
EAGLE is certified by \citet{xia2024unlocking} as the fastest speculative method in their experiments.
For each target and draft model group, we perform speculative decoding with greedy sampling and compare OPT-Tree with the Binary tree and EAGLE tree. 
\begin{figure}[t]
    \centering
    \includegraphics[width=\columnwidth]{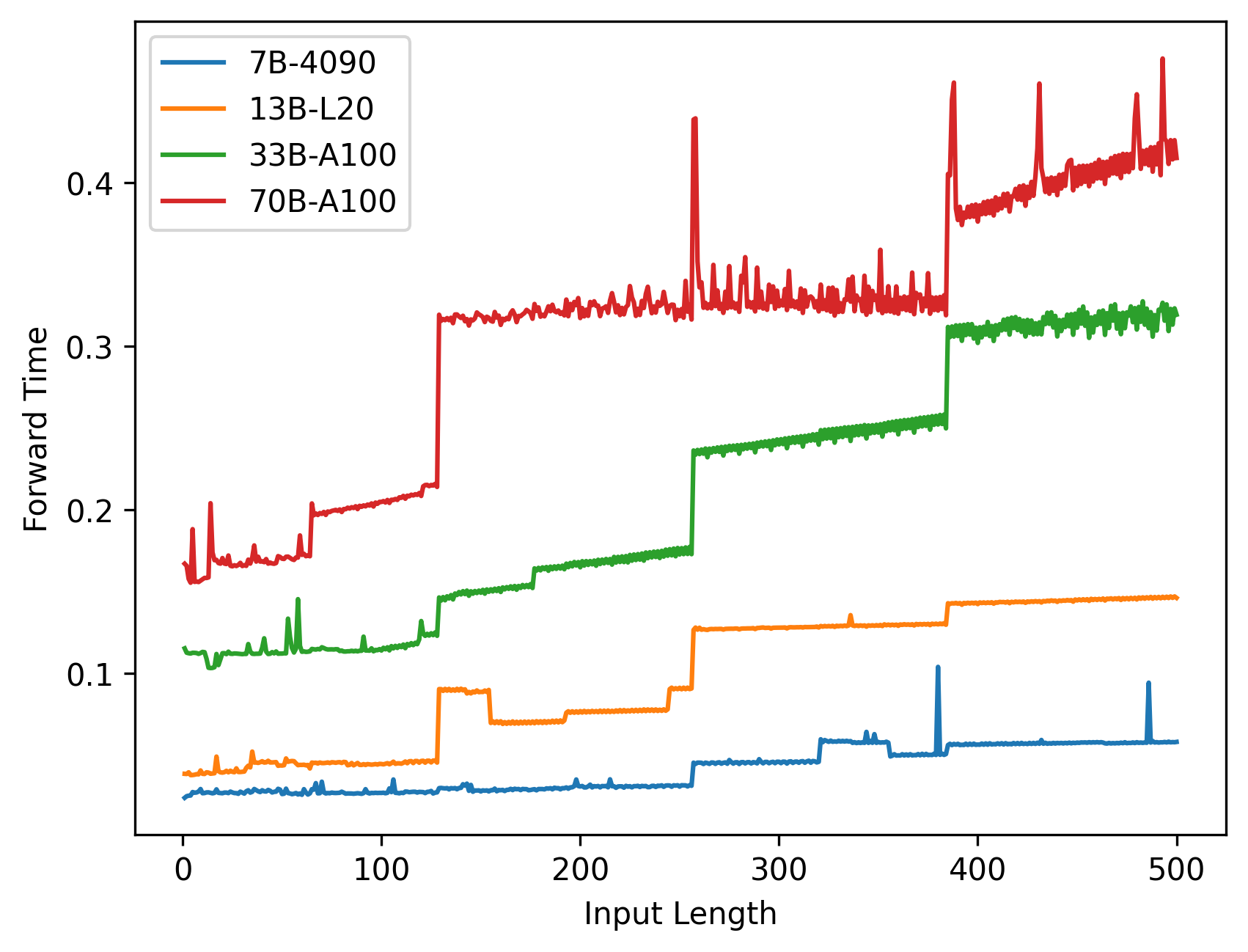}
    \caption{The relationship between input length and the wall time for inference for models of different sizes on various GPUs.}
    \label{fig:forward}
\end{figure}
\begin{table*}
\centering
\small
\renewcommand{\arraystretch}{1.45}
\resizebox{\textwidth}{!}{
\begin{tabular}{c|c|c|ccc||c|c|c|ccc} 
\toprule
\textbf{$M$}           & \textbf{$M_{d}$ }                     & Tree     & MAL    & Tokens/s & Speedup & \textbf{$M$}          & \textbf{$M_{d}$ }                     & Tree     & MAL    & Tokens/s & Speedup  \\ 
\midrule
\multirow{10}{*}{\begin{tabular}[c]{@{}c@{}}LLaMA-2\\-7B\end{tabular}} & None                    & -        & 1.00 &  52.76  & 1.00    & \multirow{10}{*}{\begin{tabular}[c]{@{}c@{}}LLaMA-2\\-13B\end{tabular}} & None                   & -        & 1.00 &     27.10    & 1.00     \\ 
\cline{2-6}\cline{8-12}
  & \multirow{3}{*}{L-68M}  & Binary   &  2.20   &  73.49     &    1.39    &                                                                        & \multirow{3}{*}{L-68M} & Binary   &  2.21    &    45.18     &   1.67   \\
                                                                      &                         & EAGLE    &   2.63   &   85.62       &    1.62     &                                                                        &                        & EAGLE    &    2.60  &  52.83        &   1.95      \\
                                                                      &                         & OPT-Tree &  \textbf{2.78}     &     \textbf{96.43}     &    \textbf{1.83}      &                                                                        &                        & OPT-Tree &   \textbf{2.81}   &       \textbf{53.54}   &    \textbf{1.98}      \\ 
\cline{2-6}\cline{8-12}
                                                                      & \multirow{3}{*}{L-1B}   & Binary   &   3.55   &    40.69   &    0.77   &                                                                        & \multirow{3}{*}{L-1B}  & Binary   & 3.76    &  36.54       &   1.35       \\
                                                                      &                         & EAGLE    &  3.87  & 44.42       &    0.84    &                                                                        &                        & EAGLE    &    4.10&    37.29   &  1.38     \\
                                                                      &                         & OPT-Tree &     \textbf{4.46} &    \textbf{50.83}     &   \textbf{0.96}     &                                                                        &                        & OPT-Tree &   \textbf{5.10}   &        \textbf{42.97}  &    \textbf{1.59}      \\ 
\cline{2-6}\cline{8-12}
                                                                      & \multirow{3}{*}{~EAGLE} & Binary   &  3.52   &     118.15   &   2.24      &                                                                        & \multirow{3}{*}{EAGLE} & Binary   & 3.80   &  73.30      &  2.70       \\
                                                                      &                         & EAGLE    &  3.83  &     137.41   &   2.60    &                                                                        &                        & EAGLE    &    4.06&   80.47     &   2.97      \\
                                                                &                         & OPT-Tree &   \textbf{4.68}   &    \textbf{140.55}     &   \textbf{2.66}      &                                                                        &                        & OPT-Tree &  \textbf{5.03}   &      \textbf{80.94}    &    \textbf{2.99}      \\ 
\midrule
\multirow{7}{*}{\begin{tabular}[c]{@{}c@{}}LLaMA-2\\-70B\end{tabular}} & None                    & -        &  1.00    &     6.38    &    1.00     & \multirow{7}{*}{\begin{tabular}[c]{@{}c@{}}Vicuna\\-33B\end{tabular}}  & None                   & -        &   1.00   &     10.74     &       1.00   \\ 
\cline{2-6}\cline{8-12}
                                                                      & \multirow{3}{*}{L-7B}   & Binary   & 4.85    &       11.20   &  1.76     &                                                                        & \multirow{3}{*}{V-7B}  & Binary   &   4.95   &  13.15        &     1.22     \\
                                                                      &                         & EAGLE    &  4.98    &     11.51     &    1.80   &                                                                        &                        & EAGLE    &   4.81   &   13.38       & 1.25         \\
                                                                      &                         & OPT-Tree &   \textbf{7.62}   &     \textbf{12.10}     &     \textbf{1.90}     &                                                                       &                        & OPT-Tree &   \textbf{6.35}  &   \textbf{13.98}       &  \textbf{1.30}        \\ 
\cline{2-6}\cline{8-12}
                                                                      & \multirow{3}{*}{EAGLE}  & Binary   &    3.62 &  18.63       &   2.92     &                                                                        & \multirow{3}{*}{EAGLE} & Binary   &  2.82    &   25.20       &   2.35       \\
                                                                      &                         & EAGLE    & 3.91    &     20.42     &     3.20    &                                                                        &                        & EAGLE    &  3.15    &   28.37       &   2.64      \\
                                                                      &                         & OPT-Tree &  \textbf{4.55}    &   \textbf{20.50}       &   \textbf{3.21}      &                                                                        &                        & OPT-Tree &   \textbf{3.47}   &   \textbf{28.76}       &   \textbf{2.68}       \\
\bottomrule
\end{tabular}
}
\caption{\label{tab:main1} Experimental results on GSM8K. $M_{d}$ being None represents vanilla autoregressive decoding. "L" and "V" in $M_{d}$ column represent "LLaMA-2" and "Vicuna". "MAL" indicates "Mean Acceptance Length".}
\end{table*}

% \begin{table}[t]
% \centering
% \small
% \resizebox{\textwidth}{!}{
% \begin{tabular}{cccccc||cccccc}
% \toprule
% M& MD & Tree & A & Tokens/s & Speedup & M & MD & Tree & A & Tokens/s & Speedup\\
%  \hline
%  \multirow{10}{*}{LLaMA-7B} & None & - & 1 & 51.89 & 1 &
%  \multirow{10}{*}{LLaMA-13B} & None & - & 1 &  & 1\\
% \hline
%  &\multirow{3}{*}{L-68M} & Binary & 0 & 0 & 0 &  &\multirow{3}{*}{L-68M} & Binary &  &  &  &\\
%  & & EAGLE &  &  &  & 
%   &  & EAGLE &  &  & \\
%  & & OPT-Tree &  &  &  & 
%    &  & OPT-Tree &  &  &\\
% \hline
%  &\multirow{3}{*}{L-68M} & Binary & 0 & 0 & 0 &  &\multirow{3}{*}{L-68M} & Binary &  &  &  &\\
%  & & EAGLE &  &  &  & 
%   &  & EAGLE &  &  & \\
%  & & OPT-Tree &  &  &  & 
%    &  & OPT-Tree &  &  &\\
% \hline
%  &\multirow{3}{*}{L-68M} & Binary & 0 & 0 & 0 &  &\multirow{3}{*}{L-68M} & Binary &  &  &  &\\
%  & & EAGLE &  &  &  & 
%   &  & EAGLE &  &  & \\
%  & & OPT-Tree &  &  &  & 
%    &  & OPT-Tree &  &  &\\
% \bottomrule
% \end{tabular}
% }
% \caption{\label{tab:main} 1}
% \end{table}
We compare the average acceptance length and number of tokens generated per second decoding with different tree structures.
The speedup ratio is calculated according to generation speed.
The node budget is determined by the target model and computational resource since the inference time generally remains the same within a certain input length.
Figure \ref{fig:forward} displays the inference time for input with various lengths for the 4 target models used in the experiments.
The number of nodes needs to be controlled within a certain range to avoid excessive time consumption in the verification phase.
It is treated as a hyperparameter chosen in $[25,50,60]$ to maximize the speedup ratio according to different target models and GPU resources except for the EAGLE tree.
We conduct evaluation on MT-Bench \citep{zheng2024judging} and GSM8K \citep{cobbe2021gsm8k}.

\noindent{\textbf{Results.}} Experimental results are shown in Table \ref{tab:main} and Table \ref{tab:main1}. 
Note that using LLaMA-2-1B as the draft model can hardly speed up decoding when the target model is LLaMA-2-7B because the difference in inference time between the two models is too small.
EAGLE draft models achieve strong performance with fewer parameters, thus providing better acceleration than the small models in the same series with the target models.
OPT-Tree outperforms other tree structures in terms of mean acceptance length in each group of experiments, especially when the performance of the draft model is close to the target model (e.g., LLaMA-2-70B combined with L-7B and Vicuna-33B combined with Vicuna-7B), indicating its high upper limit.
Since OPT-Trees are usually deeper than binary trees and EAGLE trees, they incur more overhead when drafting.
Therefore, from the perspective of tokens per second, the improvement is not as significant as that from the mean acceptance length.
Tokens per second are affected by different hardware resources and random errors.
In addition, some method-independent techniques can also be used to reduce computation time.
For example, the unchanged part of the attention mask in the drafting phase can be initialized only once and called multiple times, thus saving the time of multiple initializations.
In order to make a fairer comparison in our experiments, we avoid these tricks to be consistent with EAGLE’s practice.
Overall, OPT-Tree outperforms the baselines.
It can be up to about 3.2 times faster than vanilla autoregressive decoding.
The similar performance on both datasets verifies the robustness of the proposed method.

\subsection{Correlation between $E(A)$ and $A$}
\begin{figure}[t]
    \centering
    \includegraphics[width=\columnwidth]{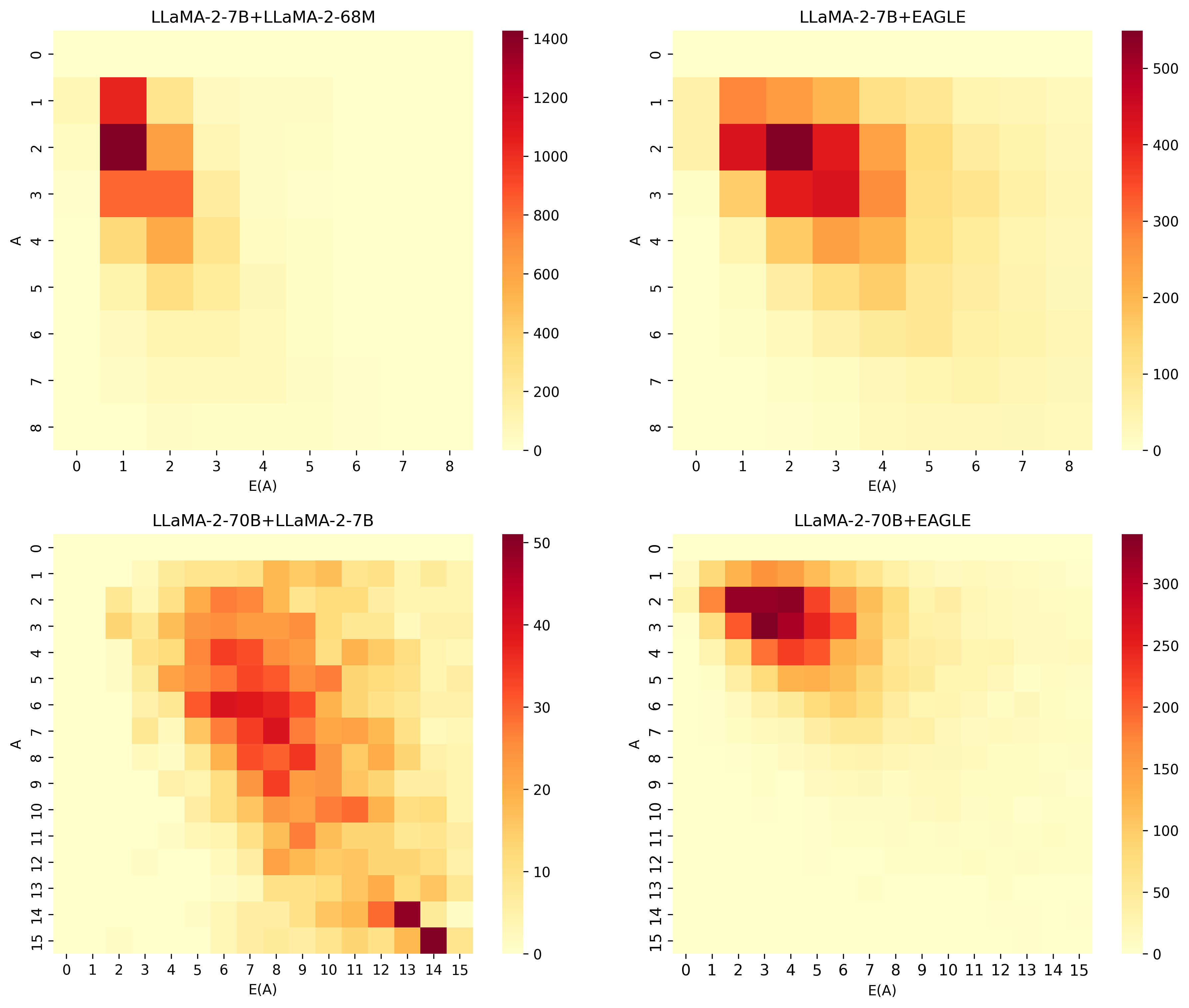}
    \caption{Correlation between $E(A)$ and $A$. The horizontal axis represents $E(A)$, and the vertical axis represents $A$. Each square shows the number of times the corresponding situation occurs. The darker the color, the more times it indicates.}
    \label{fig:e_sim}
\end{figure}
\label{sec:e}
The theory of OPT-Tree is based on the premise that $E(A)$ is positively correlated with actual $A$.
We record the values of $E(A)$ and $A$ of OPT-Tree in about 8000 decoding steps for 4 groups of $M$ and $M_{d}$.
Figure \ref{fig:e_sim} shows the results.
The value of $E(A)$ is rounded.
The darker areas in the four images are basically distributed along the main diagonal line.
When $E(A)$ of the tree is larger, it also tends to get a more considerable acceptance length after verification.
A stronger draft model shifts the distribution to the lower right corner.
These phenomena corroborate our theoretical analysis.
In addition, in the LLaMA-2-70B+LLaMA-2-7B group, high values of $E(A)$ and $A$ (e.g., $E(A)=14, A=15$) are generally found, which demonstrates the potential of OPT-Tree to adapt to stronger draft models and larger draft tree sizes.

\subsection{Scaling the Draft Tree Size}
\begin{figure}[t]
    \centering
    \includegraphics[width=\columnwidth]{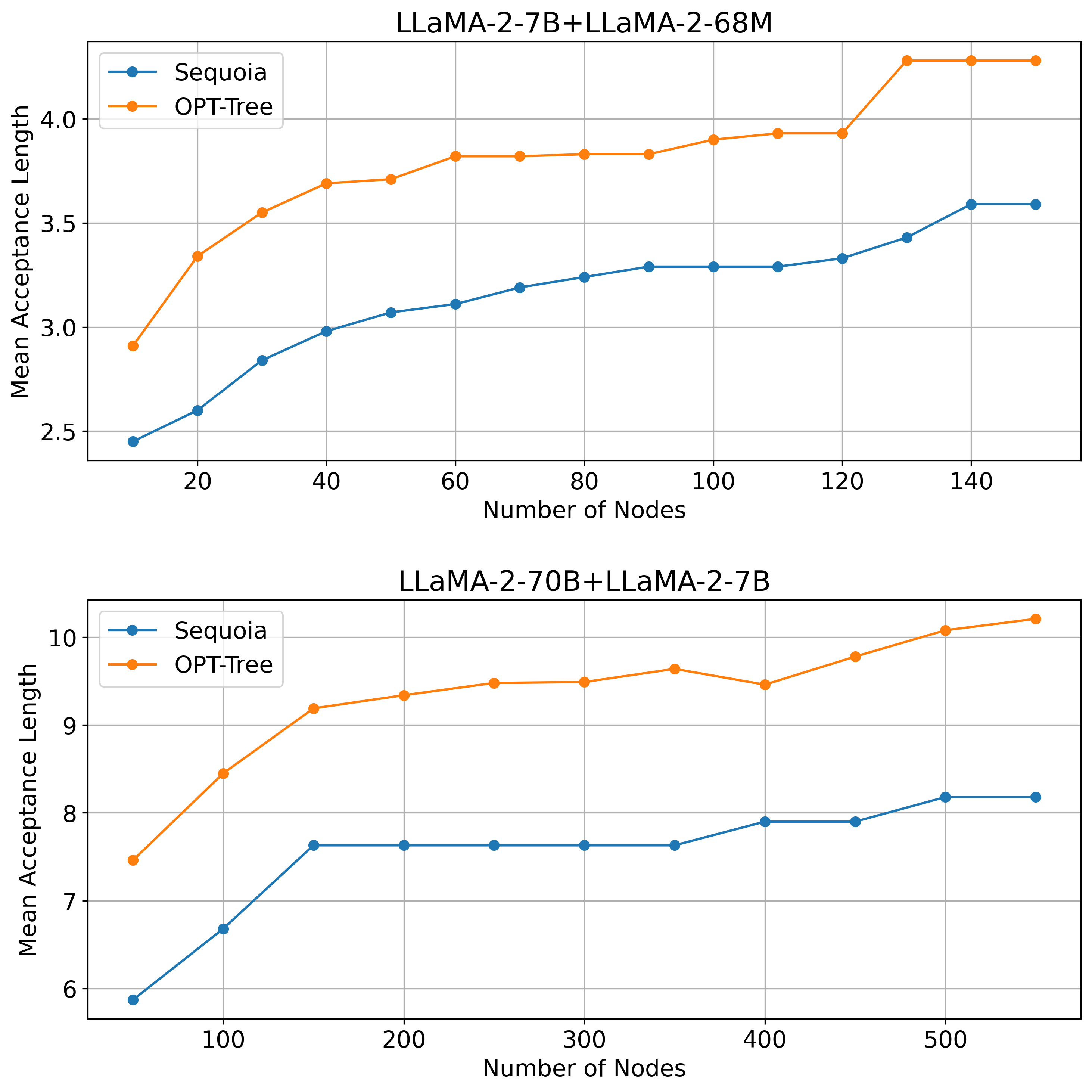}
    \caption{Mean acceptance length under different tree sizes under two sets of experiments. }
    \label{fig:scale}
\end{figure}
We conduct experiments to explore the changes in mean acceptance length with larger tree sizes.
We compare OPT-Tree with Sequoia \citep{chen2024sequoia} using LLaMA-2-7B and LLaMA-2-70B as target models.
Sequoia is a scalable draft tree that uses dynamic programming to
solve for the tree structure. 
It requires the target and draft models to be used in advance to infer some samples to determine the best structure.
The tree structure is fixed when doing speculative decoding.
We use 200 samples in C4 \citep{raffel2020exploring} to construct the Sequoia
trees.
Temperature is set to 0 in the experiments.

The experimental results are shown in Figure \ref{fig:scale}.
OPT-Tree outperforms Sequoia under various tree sizes.
For LLaMA-2-7B+LLaMA-2-68M, the mean acceptance length with both OPT-Tree and Sequoia proliferates when the number of nodes is smaller than 130.
When the number of nodes exceeds 140, the mean acceptance length increases slowly.
For LLaMA-2-70B+LLaMA-2-7B, the growth of mean acceptance length with Sequoia tends to be flat when the number of nodes exceeds 150.
However, OPT-Tree can continue to improve the mean acceptance length even if the number of nodes exceeds 500.
Since LLaMA-2-7B is a strong draft model for LLaMA-2-70B, the mean acceptance length can achieve 10 with an OPT-Tree of 500 nodes.
A tree with 500 nodes costs a large amount of computation time for LLaMA-2-70B with A100-PCIE-40GB GPUs, thus being unable to speed up decoding in our practice.
However, this cost may be acceptable if more powerful computational resources are equipped in the future.

\subsection{Impact of the Threshold}
\begin{figure}[t]
    \centering
    \includegraphics[width=\columnwidth]{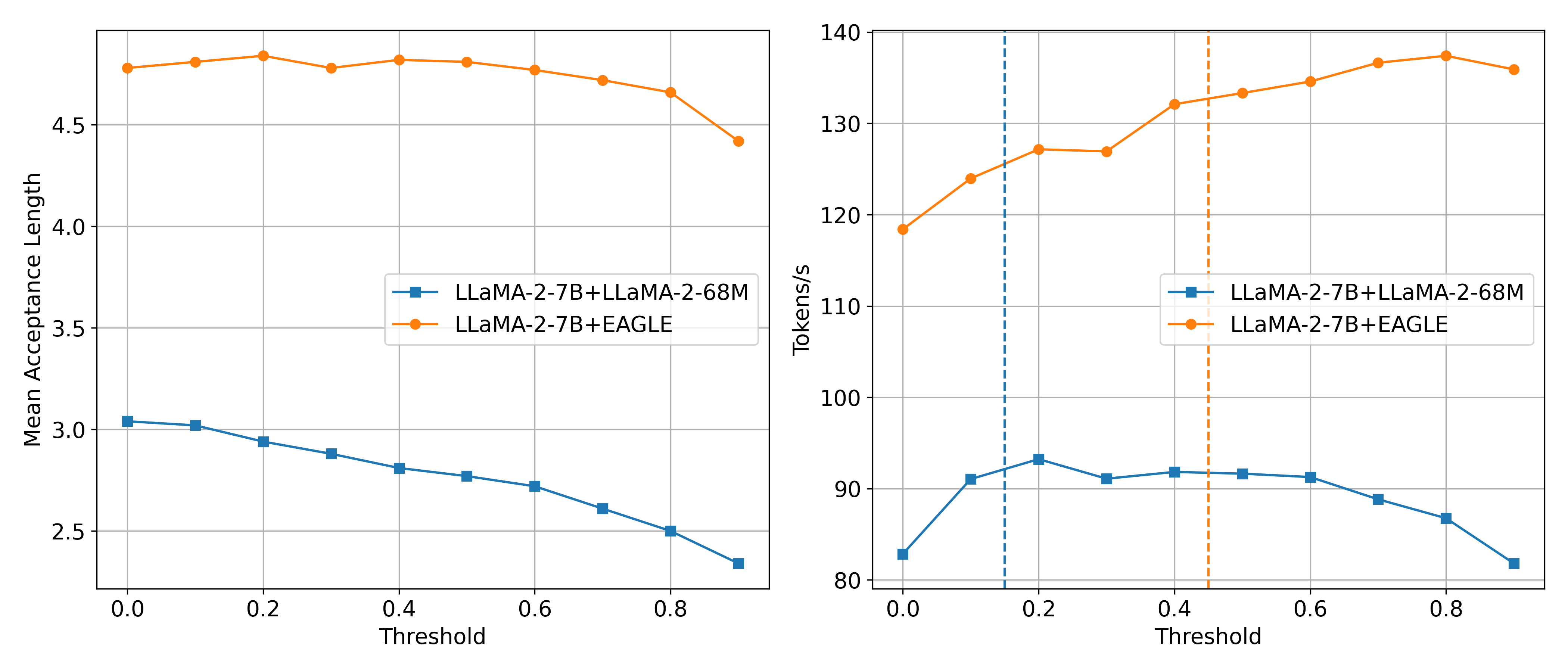}
    \caption{The two figures on the left and right are the mean acceptance length and tokens/s under different thresholds on MT-Bench. The target model is LLaMA-2-7B. The blue and orange dashed lines in the right figure represent the values of $\mu$ with LLaMA-2-68M and EAGLE as the draft model, respectively.}
    \label{fig:threshold}
\end{figure}
Considering the overhead of the draft model is proportional to the depth of the tree, the tree that maximizes the acceptance length does not necessarily have the highest speed-up ratio. Therefore, we experiment to study the mean acceptance length and tokens/s under different thresholds. 

Figure \ref{fig:threshold} shows the experimental results on LLaMA-2-7B. 
The mean acceptance length drops as the threshold grows when using LLaMA-2-68M as the draft model.
However, there is a slight fluctuation for the EAGLE draft model.
This is because $E(A)$ and $A$ are not completely equivalent.
We calculate $\mu$ for each group of models, which is the time of one drafting step divided by the time of one decoding step.
A threshold that is too large will reduce the tree's depth, thus reducing the value of $A$.
On the other hand, a threshold that is too small may make the tree too deep and increase the cost of drafting.
When the depth of the tree increases by one but the increment of the $E(A)$ does not exceed $\mu$, it is not worth increasing the depth.
So, we set a threshold between $\mu$ and 1 in practice.
LLaMA-2-68M and EAGLE achieve the highest acceleration when $\delta=0.2$ and $\delta=0.8$, respectively.

\subsection{Performance on Non-greedy Settings}
\begin{table}
\centering
\renewcommand{\arraystretch}{1.2}
\resizebox{\columnwidth}{!}{
\begin{tabular}{ccccc} 
\toprule
$M$                          & $M_{d}$ & MAL & Tokens/s & Speedup  \\ 
\midrule
\multirow{3}{*}{LLaMA-2-7B}  & L-68M   &   2.72  &    88.90      &   1.71       \\ 
% \cline{2-2}
                             & L-1B    &  5.25   &    49.76      &    0.96      \\ 
% \cline{2-2}
                             & EAGLE   &  4.07   &    125.79      &    2.42      \\ 
\hline
\multirow{3}{*}{LLaMA-2-13B} & L-68M   &   2.26  &    43.45      &    1.62      \\ 
% \cline{2-2}
                             & L-1B    &  4.23   &    37.84      &     1.41     \\ 
% \cline{2-2}
                             & EAGLE   &   4.13  &     69.27     &      2.21    \\ 
\hline
\multirow{2}{*}{LLaMA-2-70B} & L-7B    &   7.17  &     11.87     &     1.89     \\ 
% \cline{2-2}
                             & EAGLE   &  4.09   &     18.92     &     3.01     \\ 
\hline
\multirow{2}{*}{Vicuna-33B}  & V-7B    &   4.91  &     13.48     &     1.20     \\ 
% \cline{2-2}
                             & EAGLE   &   2.89  &     25.31     &      2.25    \\
\bottomrule
\end{tabular}
}
\caption{Performance of OPT-Tree on MT-Bench with the temperature set to 1. "L" and "V"
in $M_{d}$ column represents "LLaMA-2" and "Vicuna". "MAL" indicates "Mean Acceptance Length".}
\label{tab:temperature}
\end{table}
\begin{figure}[t]
    \centering
    \includegraphics[width=\columnwidth]{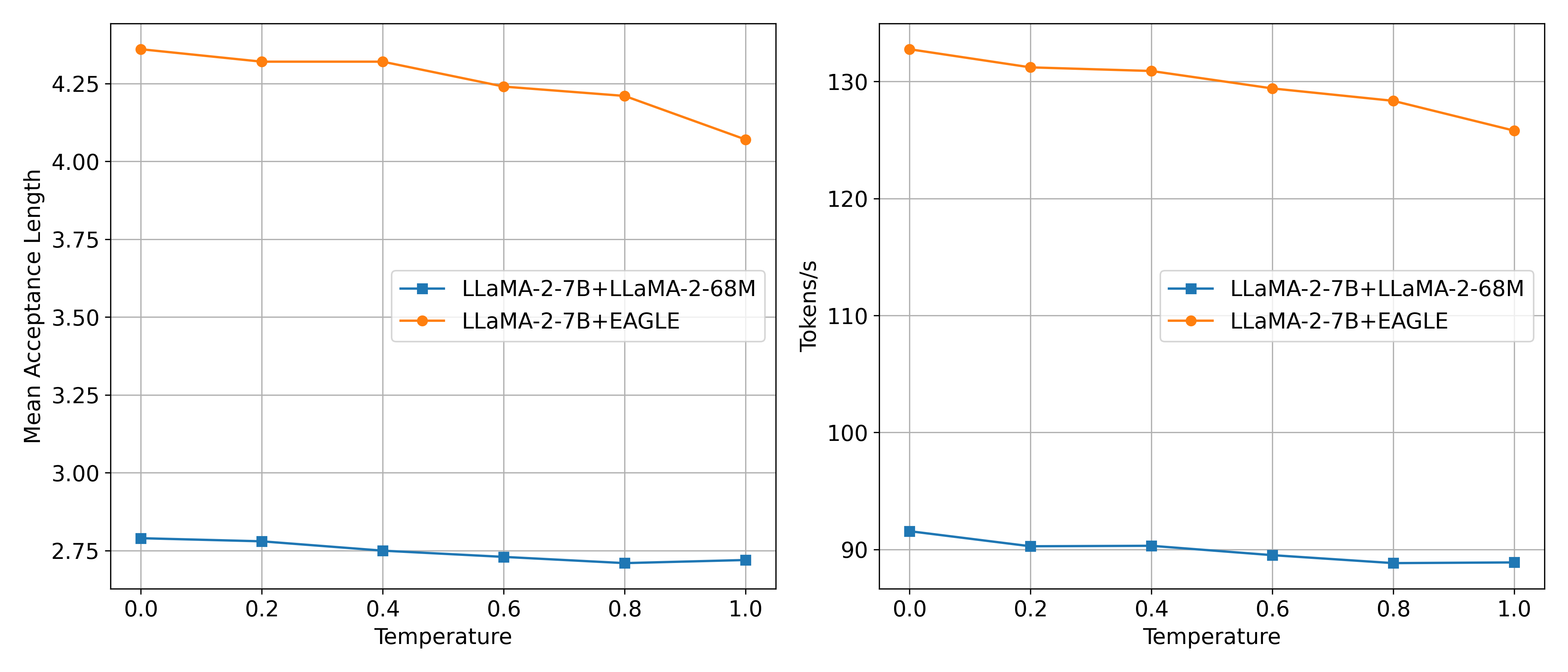}
    \caption{The two figures on the left and right are the mean acceptance length and tokens/s with OPT-Tree with different temperatures on MT-Bench. The target model is LLaMA-2-7B.}
    \label{fig:temperature}
\end{figure}
In the decoding setting of non-greedy sampling (random sampling), we only modify the acceptable tokens during the verification phase.
We conduct experiments to evaluate OPT-Tree on these non-greedy settings, where the temperature exceeds 0.

We perform speculative decoding with OPT-Tree on the MT-Bench dataset for all groups of models in \ref{sec:main} with the temperature set to 1.
Table \ref{tab:temperature} displays the experimental results.
The mean acceptance length and the speedup ratio of speculative decoding with OPT-Tree are slightly lower when the temperature is set to 1 than when the temperature is set to 0.
Since the draft tree greedily samples tokens with higher probability, the positive correlation between E(A) and A will be weakened in the decoding of random sampling.
Therefore, it is typical for the acceleration of speculative decoding to drop when the temperature is greater than 0.
Figure \ref{fig:temperature} shows specific changes in mean acceptance length and tokens/s with different temperature values.
Both metrics drop as the temperature rises in general.
But even when the temperature is set to 1, opt-tree can still provide high speedup compared to vanilla autoregressive decoding.

\subsection{Case Study}
\label{sec:case}
\begin{figure*}[t]
    \centering
    \includegraphics[width=\textwidth]{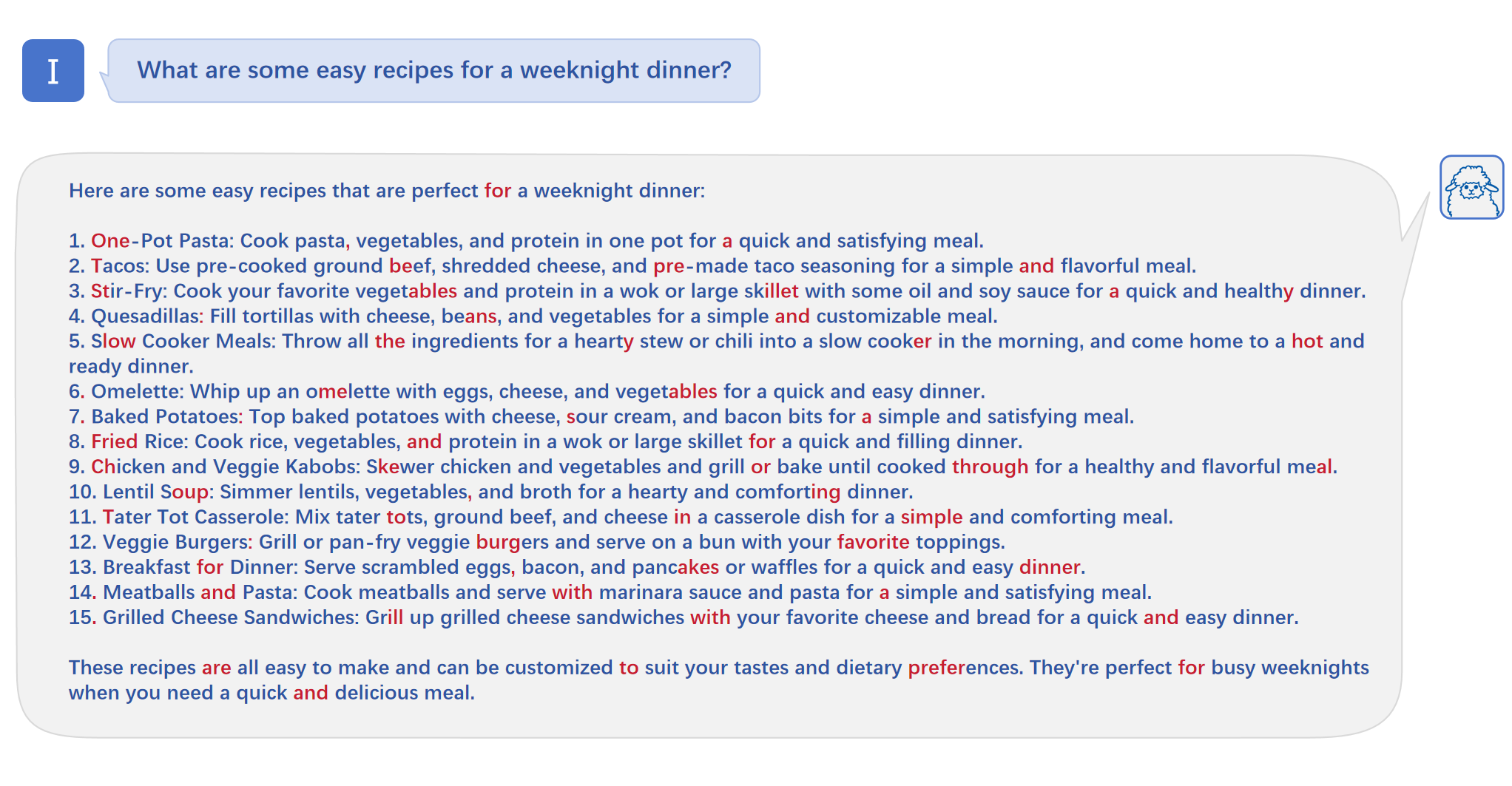}
    \caption{An example of speculative decoding with OPT-Tree on LLaMA-2-70B. Text on a blue background is the input prompt. Blue text represents drafts generated by LLaMA-2-7B and accepted by LLaMA-2-70B. Red text represents the next token for each accepted draft, which is generated by LLaMA-2-70B during the verification.}
    \label{fig:case}
\end{figure*}
We show an example of speculative decoding with an OPT-Tree of 50 nodes on LLaMA-2-70B with LLaMA-2-7B as the draft model in Figure \ref{fig:case}.
The threshold is 0.7, and the temperature is 0.
The mean acceptance length is 9.34, and the generation speed is 12.07 tokens per second.
Most words (blue text) are generated by the draft model and then verified by the target model. Each couple of red words and the continuous blue text in front of it is generated in a single decoding step of the target model.
The appearance of red words is either because the depth of the draft tree is limited or because none of the candidates for this position hits the target.
Prepositions (e.g., \textit{in}, \textit{for} and \textit{with}), conjunctions (e.g., \textit{and} and \textit{or}), articles (e.g., \textit{a} and \textit{the}), punctuation and other words which have no apparent practical meanings are prone to miss in the draft.
In addition, the beginning of new sentences in drafts tends to be rejected because it has no solid sequential association with the previous word.

\section{Related Work}
Speculative decoding \citep{stern2018blockwise,xia2023speculative,leviathan2023fast,chen2023accelerating} accelerates autoregressive decoding by drafting and then verifying while ensuring consistent output.
Drafting methods are mainly divided into independent drafting and self-drafting.
Independent drafting leverages an external low-cost model. 
SpecDec \citep{xia2023speculative} trains a non-autoregressive model for drafting while others \citep{leviathan2023fast,chen2023accelerating,spector2023accelerating,chen2023cascade,chen2024sequoia} directly utilize a smaller version of the target model.
In addition, REST \citep{he2023rest} proposed a retrieval-based drafting method.
Self-drafting uses the original information of the target model to draft. 
\citet{yang2023predictive} adopt an early-exiting mechanism for drafting.
Similarly, \citet{zhang2023draft} performs adaptive layer skipping in the drafting phase.
Lookahead Decoding \citep{fu2024break} designed an algorithm for parallel drafting and verification.
MEDUSA \citep{cai2024medusa} trains multiple decoding heads to obtain candidates for multiple steps from original features in parallel.
Considering that different sampling results at each step in drafting will affect the distribution of subsequent outputs, EAGLE \citep{li2024eagle} designed an autoregressive head, which introduced the embedding of each word in the drafting stage.

The verification method has evolved from sequence-structured verification to tree-structured verification.
Early work \citep{stern2018blockwise,leviathan2023fast,xia2023speculative,yang2023predictive,zhang2023draft,fu2024break} verifies drafts in the form of one or several sequences.
However, as the number of verification tokens increases, there are a large number of prefix duplications between sequences, resulting in redundant calculations.
To alleviate this problem, recent work \citep{he2023rest,cai2024medusa,li2024eagle,jeon2024recursive} uses heuristic tree-structured drafts and designs corresponding attention masks for parallel verification.
\citet{chen2024sequoia} proposed Sequoia, an algorithm for constructing draft trees, which performs well as the tree size scales up.

\section{Conclusion}
In this paper, we propose a novel and effective method called OPT-Tree to construct adaptive draft tree structures for speculative decoding.
OPT-Tree maximizes the mathematical expectation of the acceptance length under any limited draft tree size.
Experimental results with ten groups of target models and draft models on two datasets show that opt-tree outperforms existing draft structures.
It achieves a lossless acceleration of up to 3.2 times compared to vanilla autoregressive decoding and shows robustness on different datasets and with different temperatures.
Additionally, if equipped with a strong draft model, the mean acceptance length with OPT-Tree continues to grow even if the number of nodes is over 500, demonstrating its great potential for adapting to scenarios with more powerful computational resources.

\section*{Limitations}
Different hardware resources and environments will affect the throughput speed reported in the experiments in this article.
The experiments in this paper adopt the same decoding framework as EAGLE \citep{li2024eagle} for fair comparison.
In practice, the decoding algorithm can be optimized from other perspectives to further improve the decoding speed, which is not explored in this paper.

\bibliography{custom}

% \appendix

% \section{Example Appendix}
% \label{sec:appendix}

% This is an appendix.

\end{document}